\documentclass[letterpaper]{article} 
\usepackage{aaai25}  
\usepackage{times}  
\usepackage{helvet}  
\usepackage{courier}  
\usepackage[hyphens]{url}  
\usepackage{graphicx} 
\usepackage{natbib}  
\usepackage{caption} 
\usepackage{algorithm}
\usepackage{algorithmic}
\usepackage{amsmath}
\usepackage{xspace}
\usepackage{amsfonts}
\usepackage{bbding}

\usepackage{amssymb}  
\usepackage{booktabs} 
\usepackage{xcolor}
\usepackage{multirow}
\usepackage{subfigure}
\usepackage{pifont}
\usepackage{newfloat}
\usepackage{listings}
\DeclareCaptionStyle{ruled}{labelfont=normalfont,labelsep=colon,strut=off} 
\floatstyle{ruled}
\newfloat{listing}{tb}{lst}{}
\floatname{listing}{Listing}
\title{
SAC-ViT: Semantic-Aware Clustering Vision Transformer with Early Exit
}
\author {
    Youbing Hu\textsuperscript{\rm 1}, 
    Yun Cheng\textsuperscript{\rm 2}\thanks{Corresponding Author},
    Anqi Lu\textsuperscript{\rm 1},
    Dawei Wei\textsuperscript{\rm 3},
    Zhijun Li\textsuperscript{\rm 1}\footnotemark[1]
}
\affiliations {
    \textsuperscript{\rm 1}Faculty of Computing, Harbin Institute of Technology\\
    \textsuperscript{\rm 2}Swiss Data Science Center, Zurich, Switzerland \\
    \textsuperscript{\rm 3}School of Cyber Engineering, Xidian University\\
    youbing@stu.hit.edu.cn, yun.cheng@sdsc.ethz.ch, luanqi@stu.hit.edu.cn\\  weidawei58@gmail.com, lizhijun\_os@hit.edu.cn
}

\usepackage{bibentry}

\begin{document}

\maketitle

\begin{abstract}

The Vision Transformer (ViT) excels in global modeling but faces deployment challenges on resource-constrained devices due to the quadratic computational complexity of its attention mechanism. To address this, we propose the Semantic-Aware Clustering Vision Transformer (SAC-ViT), a non-iterative approach to enhance ViT's computational efficiency. SAC-ViT operates in two stages: Early Exit (EE) and Semantic-Aware Clustering (SAC). In the EE stage, downsampled input images are processed to extract global semantic information and generate initial inference results. If these results do not meet the EE termination criteria, the information is clustered into target and non-target tokens. In the SAC stage, target tokens are mapped back to the original image, cropped, and embedded. These target tokens are then combined with reused non-target tokens from the EE stage, and the attention mechanism is applied within each cluster. This two-stage design, with end-to-end optimization, reduces spatial redundancy and enhances computational efficiency, significantly boosting overall ViT performance. Extensive experiments demonstrate the efficacy of SAC-ViT, reducing 62\% of the FLOPs of DeiT and achieving 1.98× throughput without compromising performance.

\end{abstract}

\begin{figure}[t]
\centering
\includegraphics[width=1.0\columnwidth]{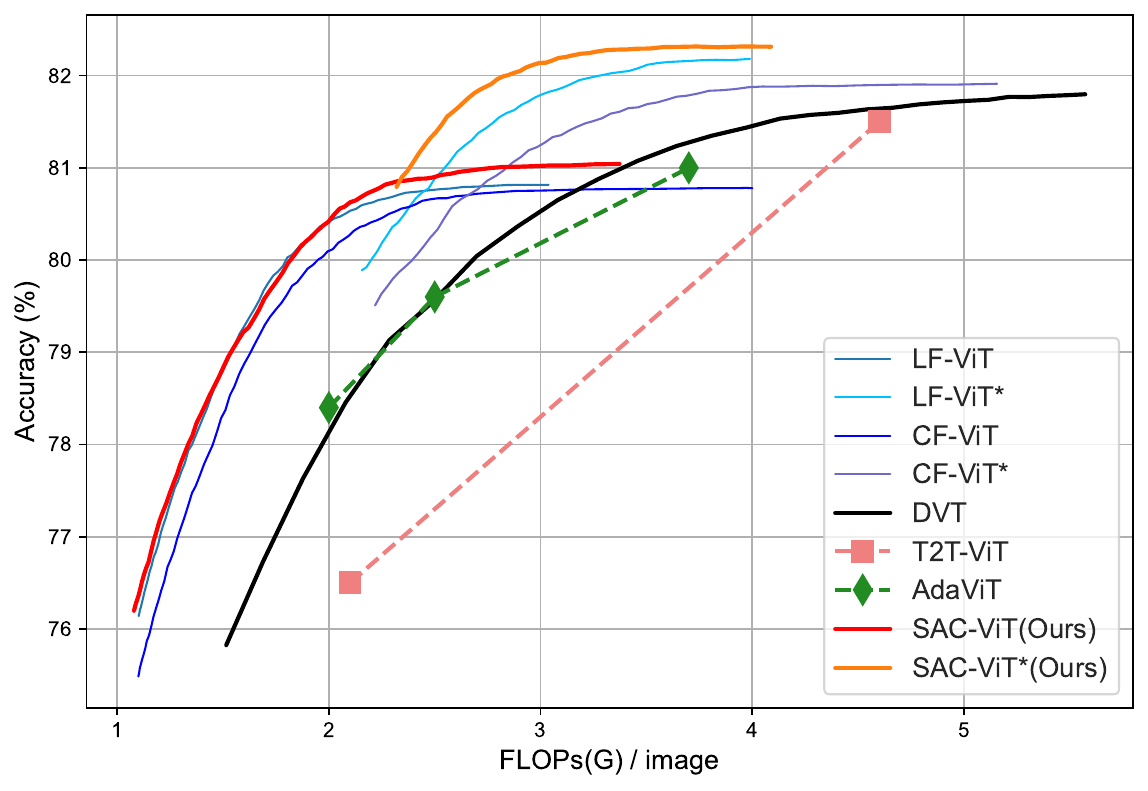}
\caption{Comparing our SAC-ViT with state-of-the-art adaptive ViT optimization methods, SAC-ViT achieves better efficiency/Top-1 accuracy trade-off. SAC-ViT, DVT \cite{wang2021not}, CF-ViT \cite{chen2023cf}, and LF-ViT \cite{hu2024lf} are all build up DeiT \cite{touvron2021training}. $*$ indicates that the input resolution is $288\times 288$.}
\label{fig:fig1}
\end{figure}

\section{Introduction}

The Vision Transformer (ViT) \cite{dosovitskiy2020image} has revolutionized computer vision tasks such as image classification \cite{he2016deep,iandola2016squeezenet,sandler2018mobilenetv2}, object detection \cite{carion2020end,zhu2020deformable,roh2021sparse}, and semantic segmentation \cite{zheng2021rethinking,strudel2021segmenter,xie2021segformer} by leveraging self-attention \cite{vaswani2017attention} to capture long-range dependencies and complex patterns. This capability has enabled ViTs to outperform traditional convolutional neural networks (CNNs) \cite{he2016deep,simonyan2014very} and achieve state-of-the-art results across various benchmarks \cite{liu2021swin,wang2022pvt}. However, the quadratic computational complexity of the attention mechanism poses significant challenges for deploying ViTs on resource-constrained devices \cite{ignatov2019ai}. As image resolution increases, so do memory and processing power demands, negatively impacting real-time processing, energy consumption, and latency. Therefore, optimizing ViT efficiency without compromising performance is crucial for broader applicability in real-world scenarios.

\begin{table*}[t]
\centering
\begin{tabular}{c|cccc}
\hline
    Method & Semantic Information & Partition Efficiency & Non-equi-partition &  Spatial Redundancy \\
    \hline
    Swin Transformer  & \ding{53} & \checkmark & \ding{53} & \ding{53}\\
    DGT & \checkmark & \ding{53} & \checkmark & \ding{53} \\
    SecViT  & \checkmark &\checkmark &  \ding{53} & \ding{53} \\
    \textbf{SAC-ViT (Ours)} & \checkmark  &  \checkmark  &  \checkmark & \checkmark \\
\hline
\end{tabular}
\caption{
Comparison among Window Partition (Swin Transformer) \cite{liu2021swin}, Dynamic Grouping by k-means (DGT) \cite{liu2022dynamic}, Semantic Equitable Clustering (SecViT) \cite{fan2024semantic}, and our SAC-ViT highlights its advantages in semantic information extraction, grouping efficiency, non-equidistant partitioning, and spatial redundancy reduction.
}
\label{tab:tab1}
\end{table*}

\begin{table}[t]
\centering
\begin{tabular}{c|cc}
\hline
    Resolutions & 224$\times$224 & 112$\times$112 \\
    \hline
    Accuracy & 79.8\% & 73.3\%  \\
    FLOPs & 4.60G & 1.10G \\
\hline
\end{tabular}
\caption{Accuracy and FLOPs of DeiT-S \cite{touvron2021training} on ImageNet \cite{deng2009imagenet} using different input image resolutions.}
\label{tab:tab2}
\end{table}

Researchers have explored various strategies \cite{chen2021crossvit,patro2023spectformer} to reduce the computational burden of the attention mechanism, including sparse attention mechanisms \cite{wang2022pvt}, low-rank approximations \cite{dass2023vitality,li2024bi}, efficient tokenization methods \cite{yin2022vit,rao2021dynamicvit}, and token grouping \cite{bolya2022tome,fan2024semantic}. Token grouping, which limits the attention span to neighboring tokens, effectively reduces the computational load for calculating attention weights and significantly improves efficiency. Due to its simplicity and effectiveness, token grouping has become a popular optimization technique in ViT.

For example, the Swin-Transformer \cite{liu2021swin} divides tokens into small windows for localized attention, while the CSWin-Transformer \cite{dong2022cswin} uses a cross-shaped grouping strategy to provide a global receptive field. MaxViT \cite{tu2022maxvit} integrates window and grid attention, allowing tokens in one window to attend to those in others. However, these approaches rely on spatial positioning and overlook semantic context, which limits the self-attention mechanism's ability to establish semantic dependencies. To address this, DGT \cite{liu2022dynamic} uses k-means clustering for query grouping based on semantic information, improving feature learning but reducing efficiency due to iterative clustering. SECViT \cite{fan2024semantic} introduces a semantically equitable clustering method, grouping tokens into equal-sized clusters in a single iteration, thereby enhancing parallel computation efficiency. Despite these advances, these methods fail to address scale and spatial redundancy, distributing computational resources uniformly across clusters and resulting in inefficiency.


In this paper, our goal is to reduce the computational cost of token partitioning methods without compromising accuracy. We observed significant spatial redundancy in images. As shown in Table~ref{tab:tab2}, training DeiT-S on the ImageNet dataset with different resolutions reveals that using a high resolution ($224\times 224$) increases accuracy by 6.5\% but also increases the computational cost by 4.2 times. This suggests that for some images, low resolution can be used for inference, allowing early exit to terminate the process sooner. However, for more complex images that require finer details, higher resolution is still necessary. Therefore, we argue that the optimal token partitioning method should adaptively cluster target and non-target areas based on the semantic content of the image, and apply the attention mechanism within each cluster separately to reduce computational costs.

To this end, we propose a non-iterative Semantic-Aware Clustering Vision Transformer (SAC-ViT) to address spatial redundancy and enhance computational efficiency. SAC-ViT operates in two stages: Early Exit (EE) and non-iterative Semantic-Aware Clustering (SAC). In the EE stage, SAC-ViT uses downsampled images to extract global semantic information and generate initial inference results. If the results meet the EE termination criteria, they are returned. Otherwise, it clusters the semantic information into target and non-target tokens. By operating on low-resolution images, SAC-ViT minimizes computational expenses, realizing efficiency gains. In the SAC stage, target tokens are mapped back to the original image, cropped, and embedded, while non-target tokens are reused to form two clusters. Attention calculations are performed within each cluster separately, optimizing computations and reducing spatial redundancy.

SAC-ViT's two stages are end-to-end optimized using the same network parameters, improving inference efficiency. As shown in Table~\ref{tab:tab1}, SAC-ViT effectively considers token semantics and spatial redundancy during clustering. Evaluations on ImageNet \cite{deng2009imagenet} using DeiT \cite{touvron2021training} demonstrate significant efficiency gains with SAC-ViT. Fig.~\ref{fig:fig1} shows the results of the comparison between SAC-ViT and state-of-the-art adaptive ViT optimization methods, indicating that SAC-ViT achieves a better efficiency/accuracy balance.

In summary, our contributions are as follows:
\begin{itemize}

    \item We propose the SAC-ViT framework, which, through a two-stage design, combines dynamic ViT compression optimization with semantic-based adaptive token grouping for the first time. This approach reduces spatial redundancy in non-target areas, thereby enhancing ViT compression and overall efficiency.
    
    \item We conduct comprehensive and rigorous experiments on the ImageNet dataset, demonstrating that SAC-ViT significantly improves efficiency compared to state-of-the-art methods. For instance, SAC-ViT reduces 62\% of the FLOPs of DeiT and achieves 1.98× throughput without compromising performance.
\end{itemize}

\section{Related Work}

\subsection{Vision Transformer}
The Vision Transformer (ViT) \cite{dosovitskiy2020image}, since its inception, has garnered significant attention from the vision community \cite{liu2021swin,wang2022pvt,xie2021segformer,carion2020end,ding2022davit,han2021transformer} for its superior global modeling capabilities. To optimize ViT's training and inference speed, various methods have been proposed \cite{dong2022cswin,fan2023rethinking,fan2024lightweight,ren2023sg,touvron2021training}. For instance, DeiT \cite{touvron2021training} uses a distillation token to transfer knowledge from a pre-trained teacher model to a student model, enhancing performance and accuracy. LV-ViT \cite{jiang2021all} leverages all tokens to compute the training loss, with location-specific supervision for each patch token. Additionally, some methods enhance ViT's architecture, such as CPVT \cite{chu2021conditional}, which replaces learnable positional embedding with a convolution layer, and CaiT \cite{touvron2021going}, which builds deeper transformers with specialized training strategies. TNT \cite{han2021transformer} models pixel-wise interactions within each patch using an inner block, preserving richer local features. Token downsampling methods have also been employed, such as PVT \cite{wang2021pyramid}, which uses average pooling. CMT \cite{guo2022cmt} and PVTv2 \cite{wang2022pvt}, which combine downsampling with convolution to maintain feature integrity. STViT \cite{chang2023making} captures global dependencies by sampling super tokens, applying self-attention, and mapping them back to the original token space for efficient global context modeling. 

\subsection{ViT Optimisation}

\textbf{ViT Compression Optimisation.} Existing research on ViT compression can be classified into static and dynamic categories based on input dependency \cite{liu2021swin,wang2022pvt,wang2021pyramid,rao2021dynamicvit,chen2023cf}. Static ViT compression involves efficient architectures like hierarchical Transformers \cite{liu2021swin,wang2022pvt} and hybrid models combining CNNs and ViTs \cite{hatamizadeh2023fastervit,zhao2022lightweight}. Some methods replace global self-attention with local self-attention \cite{liu2021swin} to reduce computational costs. Dynamic ViT compression adjusts the computational graph based on the input, dynamically removing non-contributory tokens during inference \cite{meng2022adavit,yin2022vit,rao2021dynamicvit,pan2021ia,liang2022not} or allocating computational resources to different image regions based on their significance \cite{xu2022evo,wang2021not,chen2023cf,hu2024lf,tang2022quadtree,tang2022patch}. Our SAC-ViT also allocates resources to target and non-target tokens based on semantic information, performing local self-attention separately in these tokens to reduce computation costs.

\textbf{Grouping-Based ViT Optimisation.} Token grouping \cite{ding2022davit,dong2022cswin,liu2022dynamic,liu2021swin,tu2022maxvit,bolya2022tome} optimizes ViT by limiting each token's attention span to neighboring tokens, reducing computational load. The Swin Transformer \cite{liu2021swin} divides tokens into small windows for localized attention, while the CSWin-Transformer \cite{dong2022cswin} uses a cross-shaped grouping strategy for a global receptive field. MaxViT \cite{tu2022maxvit} combines window and grid attention mechanisms. However, these methods often overlook the semantic context of tokens, limiting their ability to capture semantic dependencies. To address this, the Dynamic Grouping Transformer (DGT) \cite{liu2022dynamic} uses k-means clustering for query grouping, incorporating semantic information to enhance feature learning. The Semantic Equitable Clustering Vision Transformer (SecViT) \cite{fan2024semantic} groups tokens into equal-sized clusters in a single iteration, enhancing parallel computation efficiency. Despite these advancements, existing methods often ignore targets' scale and spatial redundancy, leading to inefficiencies as identical resources are allocated to all clusters.
In contrast, our SAC-ViT improves ViT compression and overall efficiency by reducing spatial redundancy in non-target regions through a two-stage design that, for the first time, combines dynamic ViT compression optimization with adaptive token grouping based on semantic information.

\begin{figure*}[t]
\centering
\includegraphics[width=1.9\columnwidth]{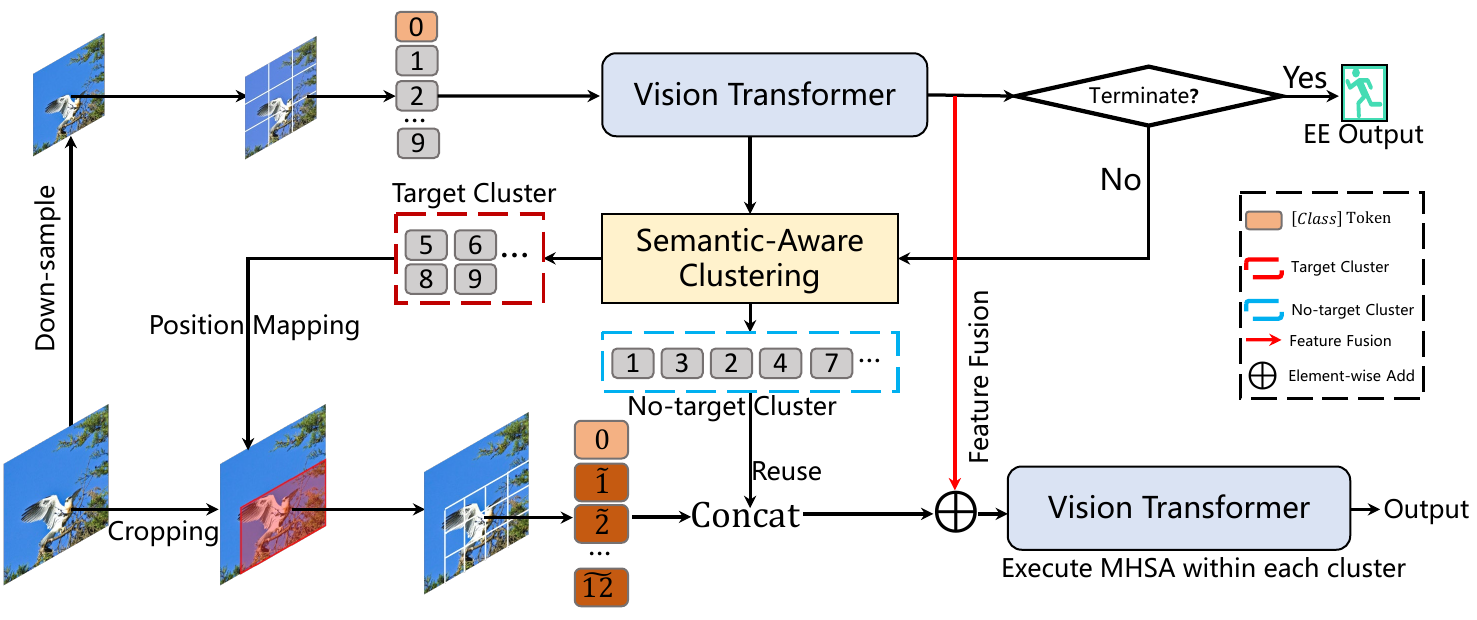}
\caption{
Overview of the SAC-ViT framework. SAC-ViT consists of an Early Exit (EE) stage and a non-iterative Semantic-Aware Clustering (SAC) stage. In the EE stage, downsampled images are processed to extract global semantic information and generate initial results. If these results don't meet the EE terminate criteria, the information is clustered into target and non-target tokens. In the SAC stage, target tokens are mapped back to the original image, cropped, embedded, and then combined with reused non-target tokens from the EE stage. Multi-Head Self-Attention (MHSA) is applied within each cluster. Notably, SAC-ViT uses the same network parameters in both stages and performs end-to-end optimization.
}
\label{fig:sac_vit}
\end{figure*}

\section{Preliminaries}
Vision Transformers (ViTs) \cite{dosovitskiy2020image} process images by dividing an input image \(\mathbf{I} \in \mathbb{R}^{H \times W \times C}\) into fixed-size patches of \(P \times P\) pixels, resulting in \(N = \frac{H \times W}{P^2}\) patches. Each patch is then flattened and linearly projected to form patch embeddings:
\begin{equation}
    \mathbf{x} = [\mathbf{x}_1; \mathbf{x}_2; \ldots; \mathbf{x}_N] \in \mathbb{R}^{N \times D},
\end{equation}
where \(\mathbf{x}_i \in \mathbb{R}^{D}\) represents the embedded patch vector. 
To retain spatial information, positional encodings \(\mathbf{P} \in \mathbb{R}^{(N+1) \times D}\) are added to the patch embeddings. Additionally, a classification token \(\mathbf{x}_{\text{cls}} \in \mathbb{R}^{D}\) is prepended to the sequence of embedded patches, resulting in:
\begin{equation}
\mathbf{Z}_0 = [\mathbf{x}_{\text{cls}}; \mathbf{x}] + \mathbf{P}.
\end{equation}
The self-attention mechanism then processes these embeddings. For each embedding \(\mathbf{Z} \in \mathbb{R}^{(N+1) \times D}\), query (\(\mathbf{Q}\)), key (\(\mathbf{K}\)), and value (\(\mathbf{V}\)) matrices are computed:
\begin{equation}
\mathbf{Q} = \mathbf{Z} \mathbf{W}_Q, \quad \mathbf{K} = \mathbf{Z} \mathbf{W}_K, \quad \mathbf{V} = \mathbf{Z} \mathbf{W}_V,
\end{equation}
where $ \mathbf{W}_Q, \mathbf{W}_K,$ and $ \mathbf{W}_V \in \mathbb{R}^{D \times D}$ are learned projection matrices. The attention scores are then calculated and normalized:
\begin{equation}
\mathbf{A} = \text{Softmax}\left(\frac{\mathbf{Q} \mathbf{K}^\top}{\sqrt{D}}\right) = [\mathbf{a}_{cls}; \mathbf{a}_1; \ldots; \mathbf{a}_N].
\end{equation}
The class attention \(\mathbf{a}_{cls}\) represents the entire image. The self-attention mechanism's output is a weighted sum of the values: \(\mathbf{Z}' = \mathbf{A} \mathbf{V}\). Multi-head self-attention (MHSA) captures different input aspects by performing parallel self-attention operations. The outputs from all heads are then concatenated and linearly transformed.
\begin{equation}
\mathbf{Z}'_{\text{MH}} = \text{concat}(\mathbf{Z}'_1, \mathbf{Z}'_2, \ldots, \mathbf{Z}'_h) \mathbf{W}_O,
\end{equation}
where \(\mathbf{W}_O \in \mathbb{R}^{h \times D \times D}\) is the output projection matrix. The output embeddings are then processed by a feed-forward network (FFN). Layer normalization (LN) and residual connections are applied to stabilize and accelerate training.
\begin{equation}
\mathbf{Z}_{\text{out}} = \text{LN}(\mathbf{Z} + \text{FFN}(\text{LN}(\mathbf{Z} + \mathbf{Z}'_{\text{MH}}))).
\end{equation}
The output \( Z_{out}^L \) after \( L \) layers of MHSA-FFA transformations is fed into the classifier to obtain the final classification.

In the MHSA-FFN transformation, the computational complexities of MHSA and FFN \cite{liu2021swin} are:
\begin{equation}
\label{eq7}
\begin{aligned}
O(MHSA) = 4ND^2 + 2N^2D, \\
  O(FFN) = 8ND^2.
\end{aligned}
\end{equation}
We can see that ViT's computational complexity is quadratic in both the embedding dimension \(D\) and the number of tokens \(N\). Since the embedding dimension \(D\) is fixed, reducing the number of input tokens \cite{yin2022vit,chen2023cf,hu2024lf,rao2021dynamicvit} effectively lowers the ViT's complexity, which is also the focus of this paper.

\section{Semantic-Aware Clustering Vision Transformer with Early Exit}
In this section, we formally introduce our non-iterative Semantic-Aware Clustering Vision Transformer (SAC-ViT) with early exit to optimize ViT efficiency. We aim to: (1) reduce the number of $N$ in Eq.~\ref{eq7} by minimizing spatial redundancy, and (2) lower computational complexity from \(N^2\) to \((N-M)^2 + M^2\) by applying clustering-aware local self-attention to target and non-target tokens, where \(M\) is the number of target tokens. As shown in Fig.~\ref{fig:sac_vit}, the downsampled low-resolution image is input into SAC-ViT to extract global semantic information and generate initial inference results. If these results do not meet the EE termination criteria, the semantic information is clustered into target and non-target tokens, followed by the semantic-aware clustering stage. Details are provided below.

\subsection{Early Exit Stage}
SAC-ViT begins with an early exit (EE) stage to extract global semantic information. Specifically, for an input image \(\mathbf{I} \in \mathbb{R}^{H \times W \times C}\), SAC-ViT first downsamples it to \(\Tilde{\mathbf{I}} \in \mathbb{R}^{\frac{H}{2} \times \frac{W}{2} \times C}\). This downsampled image is then input into a ViT network with $L$ encoders to extract global features. Each encoder consists of a multi-head self-attention (MHSA) mechanism and a feed-forward network (FFN). 
The classification tokens \(z_{cls}^L\) output by the \(L\)-th encoder are fed into a classifier \(\mathcal{F}\) to obtain the prediction distribution \(\mathbf{p}\):
\begin{equation}
    \mathbf{p} = \mathcal{F}(z_{cls}^L) = [p_1, p_2, \cdots, p_C],
\end{equation}
where \(C\) denotes the number of categories in the classification task. The class \(j\) with the highest probability in the distribution is taken as the predicted classification:
\begin{equation}
    j = \mathop{\arg\max}\limits_{i} p_i.
\end{equation}

To reduce the computational cost of ViT, we introduce an early exit strategy \cite{kaya2019shallow,han2021dynamic} to terminate the inference of easily recognizable samples. Specifically, we compare the probability \(p_j\) of the predicted class \(j\) with a threshold \(\eta\). If \(p_j > \eta\), we immediately terminate the inference and return \(j\) as the final result. Otherwise, we proceed with semantic-aware clustering to distinguish between target tokens and non-target tokens.

\subsection{Semantic-Aware Clustering Stage}

\textbf{Semantic-Aware Clustering.} Previous research \cite{chen2023cf,lin2023super,hu2024lf} has shown that the attention score \(a_{cls}\) can serve as an indicator of token importance. Inspired by this, we cluster tokens in the EE stage into target and non-target tokens based on their attention scores. Specifically, we use the global moving average of attention scores across the entire network as the average score for each token:
\begin{equation}
    \mathbf{\overline{a}}_{cls}^l = \beta \cdot \mathbf{\overline{a}}_{cls}^{l-1} + (1 - \beta) \cdot \mathbf{a}_{cls}^l,
\end{equation}
where \(\beta = 0.99\). After obtaining the score of each token, we sort these scores and introduce a tunable hyperparameter \(\alpha\) to cluster the top-\(M\) tokens as the target token set $T$ and the remaining \((N - M)\) tokens as the non-target (background) token set $T^\prime$, where \(M = \lfloor \alpha N \rfloor\).

Although both our method and previous methods \cite{chen2023cf,lin2023super,hu2024lf} use attention scores to indicate token importance, our EE stage clusters tokens into target and non-target groups, while previous methods only identify token importance. Compared to iterative clustering methods \cite{liu2022dynamic,bolya2022tome,fan2024semantic}, our approach is non-iterative, and leverages global semantic information and early exit termination strategies, making it more efficient.

Once the \(M\) target tokens are identified, their locations are mapped from \(\Tilde{\mathbf{I}}\) to \(\mathbf{I}\). We then crop these regions and embed them to achieve a high-resolution tokenized representation of the target tokens. To expedite this process, we embed the entire original image and select tokens based on indices. Each target token is represented by four tokens in the original image, resulting in \(4M\) target tokens. For each token in the target token set \(T\) with index \(i\), the top-left token after mapping is represented as \(id_1\). The mapped representation of the token with index \(i\) is then:
\begin{equation}
    (id_1, id_1 + 1, id_1 + 2H_1, id_1 + 2H_1 + 1),
\end{equation}
where \(id_1 = 4i - 2(i \% H_1)\), \(H_1 = \lfloor H/2P \rfloor\), and \(P\) is the patch size.

We map the target token set \(T\) to the set \(\tilde{T}\), representing all tokens in the original image. To minimize spatial redundancy, we reuse the non-target background token set \(T^\prime\) from the EE stage, input \(\tilde{T}\) and \(T^\prime\) together into the ViT for local MHSA computation. Specifically, for the target token set \(\tilde{T}\) and the non-target token set \(T^\prime\), MHSA is performed within each cluster in every encoder, followed by using FFN to integrate the channel information of the tokens. Note that the class token is shared and updated between the two clusters. The bottom of Fig.~\ref{fig:sac_vit} shows the whole SAC stage.

\textbf{Feature Fusion.} Feature fusion \cite{mungoli2023adaptive,dai2021attentional} is a widely adopted technique to enhance feature representation capabilities. Similar to previous works \cite{chen2023cf,hu2024lf}, we aim to further improve the feature representation in the SAC stage by performing feature fusion between the target tokens from the EE stage and the mapped target tokens from the SAC stage. Since the number of target tokens in the SAC stage is four times that of the EE stage, we first use an MLP to map the number of tokens in set \(T\) to match that in set \(\tilde{T}\). Then, we perform token-level addition to achieve feature fusion.

\textbf{SAC-ViT Computational Complexity Analysis.}  SAC-ViT reduces the number of tokens by a factor of 4 in the EE stage compared to the general ViT model's MHSA. In the SAC phase, local MHSA computations are performed within each cluster. Consequently, SAC-ViT optimizes the computational complexity of MHSA in Eq.~\ref{eq7} to Eq.~\ref{eq8}:
\begin{equation}
\label{eq8}
    \begin{aligned}
        O(\text{MHSA}) &= \underbrace{4ND^2 + 2N^2D}_{\text{EE stage}} 
        + \underbrace{16MD^2 + 2(4M)^2D}_{\text{SAC stage target tokens}} \\
        &\quad + \underbrace{4(N-M)D^2 + 2(N-M)^2D}_{\text{SAC stage non-target tokens}}.
    \end{aligned}
\end{equation}
Here, \(N\) represents the total number of tokens in the EE stage, while \(M\) denotes the number of target tokens, with the constraint that \(M < N\).

Since SAC-ViT incorporates an early exit strategy at the EE stage, as indicated by the experimental results presented in Fig.~1 within the supplementary material, most samples are correctly identified at this stage. Only a small number of difficult samples proceed to the SAC computation. This process significantly reduces the spatial redundancy of the images, leading to a considerable decrease in the number of tokens. Consequently, the computational load is drastically lowered, enhancing SAC-ViT's efficiency while maintaining robust feature representation capabilities. As shown in Eq.~\ref{eq8}, the effect of our method becomes more pronounced with larger $N$, which is further verified by the results in Fig.~\ref{fig:fig1} for $288 \times 288$ input resolution.

\subsection{Training Objective}

The objective of SAC-ViT is to maximize the accurate recognition of samples during the EE stage while minimizing computational cost and ensuring accurate identification of all samples during the SAC stage. To achieve this, we adopt a two-step supervision approach: soft labels guide training in the EE stage, and ground truth (GT) labels are used for supervision in the SAC stage, following established methodologies \cite{chen2023cf,lin2023super,hu2024lf}. The loss function is defined as:

\begin{equation}
\label{eq9}
    \mathcal{L}_{loss} = CE(\mathbf{p}_{\text{sac}}, \mathbf{y}) + KL(\mathbf{p}_{\text{ee}}, \mathbf{p}_{\text{sac}}),
\end{equation}
where \( CE(\cdot, \cdot) \) and \( KL(\cdot, \cdot) \) represent the cross-entropy loss and Kullback-Leibler divergence, respectively. \( \mathbf{p}_{\text{ee}} \), \( \mathbf{p}_{\text{sac}} \), and \( \mathbf{y} \) denote the outputs of the EE stage, the outputs of the SAC stage, and the ground truth labels, respectively.

During training, it is crucial to set the threshold \( \eta \) to 1 to ensure that all samples participate in SAC-ViT's two-stage training. During inference, adjusting \( \eta \) allows for models with varying accuracy and computational costs. A higher \( \eta \) value results in more samples proceeding to the SAC stage, increasing computational costs but improving accuracy. Conversely, a lower \( \eta \) value causes more samples to exit early in the EE stage, reducing computational costs at the expense of accuracy.

\section{Experiments}
\subsection{Implementation Details}
We evaluate our SAC-ViT model on the ImageNet-1K \cite{deng2009imagenet} dataset, building it upon the Deit-S \cite{touvron2021training} model. The patch size of SAC-ViT is uniformly set to $16 \times 16$, and the default parameter $\alpha$ is 0.5. To compare under similar computational costs (FLOPs) and demonstrate the advantages of SAC-ViT, we implement our method at resolutions of $224 \times 224$ and $288 \times 288$, referred to as SAC-ViT and SAC-ViT*, respectively. For SAC-ViT, the input image resolution during the EE stage is $112 \times 112$, resulting in 49 tokens. For SAC-ViT*, the input image resolution during the EE stage is $144 \times 144$, resulting in 81 tokens.

All training settings for SAC-ViT, including learning rate, data augmentation strategies, optimizer, and regularization, strictly follow the default settings of Deit \cite{touvron2021training}. However, SAC-ViT requires a longer training time, so we use 350 epochs by default. To improve training speed and accelerate convergence, we do not use semantic-aware clustering during the first 200 epochs, instead performing SAC stage calculations on the tokens of the entire image. Semantic-aware clustering optimization is enabled only in the remaining epochs, where tokens from the EE stage are reused for non-target tokens, significantly reducing the number of tokens. Our SAC-ViT training is conducted on a workstation equipped with 10 NVIDIA 4090 GPUs. SAC-ViT always shares the same network parameters in both stages, including the feature extractor encoder, positional embedding, and classifier.

\subsection{Experimental Results}
\begin{table}[t]\small
\centering
\begin{tabular}{ccccc}
\hline
    \multirow{2}{*}{  Model}  &\multirow{2}{*}{$\eta$} & Top-1 Acc. & FLOPs & Throughput  \\
     &  & (\%) & (G) & (img./s) \\
    \hline
    DeiT-S & - &79.8 &4.63 & 2601  \\
    \hline
    SAC-ViT  &  0.45 &79.8\color{blue}{(+0.0)} & 1.75 \color{blue}{($\downarrow$62\%)} & 5140\color{blue}{($\uparrow$1.98$\times$)} \\
    SAC-ViT  &  0.47 &80.0\color{blue}{(+0.2)} & 1.81 \color{blue}{($\downarrow$61\%)} & 5055\color{blue}{($\uparrow$1.94$\times$)} \\
    SAC-ViT  &  0.52 &80.5\color{blue}{(+0.7)} & 2.03 \color{blue}{($\downarrow$56\%)} & 4813 \color{blue}{($\uparrow$1.85$\times$)}\\
    SAC-ViT  &  0.62 &80.8\color{blue}{(+1.0)} & 2.21 \color{blue}{($\downarrow$52\%)} & 4386\color{blue}{($\uparrow$1.69$\times$)}\\
    SAC-ViT  &  0.74 &81.0\color{blue}{(+1.2)} & 2.61 \color{blue}{($\downarrow$44\%)} & 3722\color{blue}{($\uparrow$1.43$\times$)}\\
    SAC-ViT  &  0.85 &81.1\color{blue}{(+1.3)} & 3.52 \color{blue}{($\downarrow$24\%)} & 2844\color{blue}{($\uparrow$1.09$\times$)}\\
    \hline
\end{tabular}
\caption{Comparison between SAC-ViT and its backbones. }
\label{tab:tab3}
\end{table}

\textbf{Comparison with backbone model.} To illustrate the efficiency of SAC-ViT, we first compare it with the base model. Consistent with previous approaches \cite{chen2023cf,hu2024lf}, our evaluation metrics include the model's Top-1 accuracy, computational costs (FLOPs), and throughput. Throughput is defined as the number of images processed per second on a single NVIDIA 4090 GPU. Specifically, we use a validation set of 50,000 images from ImageNet, processed in batches of 1024. We record the total inference time, \(T\), and compute throughput as \(50,000 / T\).

Table~\ref{tab:tab3} presents the results of SAC-ViT at different thresholds. The experimental results indicate that SAC-ViT outperforms DeiT-S in terms of computational efficiency and throughput while maintaining or improving accuracy. At the lowest threshold ($\eta = 0.45$), SAC-ViT achieves the same Top-1 accuracy (79.8\%) as DeiT-S but with 62\% fewer FLOPs and nearly double the throughput (5140 images/second). As the threshold increases, SAC-ViT’s Top-1 accuracy improves, reaching up to 81.1\% at $\eta$ = 0.85. Although higher thresholds result in increased FLOPs and slightly reduced throughput, SAC-ViT still maintains lower computational costs and higher throughput compared to DeiT-S. This makes SAC-ViT a versatile and efficient alternative to DeiT-S across various operational conditions.

\begin{table}[t]\small
\centering
\begin{tabular}{ccc}
\hline
    Method &  Top-1 Acc.(\%) & FLOPs(G) \\
    \hline
    Baseline\cite{touvron2021training} & 79.8 & 4.6 \\
    DynamicViT\cite{rao2021dynamicvit} & 79.3 & 2.9 \\
    IA-RED$^{2}$  \cite{pan2021ia} & 79.1 & 3.2 \\
    PS-ViT\cite{tang2022patch} &79.4 & 2.6 \\
    EViT \cite{liang2022not} &  79.5&3.0 \\
    Evo-ViT\cite{xu2022evo} & 79.4& 3.0 \\
    A-ViT-S \cite{yin2022vit} & 78.6& 3.6 \\
    PVT-S\cite{wang2021pyramid} & 79.8 & 3.8 \\
    SaiT-S \cite{li2022sait} & 79.4 & 2.6 \\
    Swin-T \cite{dong2022cswin} & 81.3 & 4.5 \\
    SecViT \cite{fan2024semantic} & 78.6 & 3.6 \\
    CF-ViT\cite{chen2023cf} & 80.8& 4.0 \\
    CF-ViT$^*$\cite{chen2023cf} & 81.9& 4.8 \\
    LF-ViT\cite{hu2024lf} & 80.8 & 2.4 \\
    LF-ViT*\cite{hu2024lf} & 82.2 & 3.7 \\
    \hline
    \textbf{SAC-ViT($\eta = 0.45$)} & \textbf{79.8} & \textbf{1.8} \\
    \textbf{SAC-ViT($\eta = 0.62$)} & \textbf{80.8} & \textbf{2.2} \\
    \textbf{SAC-ViT$^*$($\eta = 0.73$)} & \textbf{82.3} & \textbf{3.2} \\
\hline
\end{tabular}
\caption{Comparisons between existing token optimization methods and our SAC-ViT. $*$ indicates that the input resolution is $288\times 288$. To ensure a fair comparison, we reconstruct SecViT \cite{fan2024semantic} based on the same base model Deit-S \cite{touvron2021training}.}
\label{tab:tab4}
\end{table}

\begin{table}[t]
\centering
\begin{tabular}{ccc}
\hline
     \multirow{2}{*}{  Methods}  & \multicolumn{2}{c}{ Top-1 Acc.(\%)}\\
     & early exit & SAC \\
\hline
Naïve cluster & 65.4 & 78.5\\
Random cluster & 74.8 & 79.9\\
    \hline
    \textbf{SAC-ViT (Ours)} & \textbf{76.2} & \textbf{81.1} \\
    \hline
\end{tabular}
\caption{Comparison between SAC-ViT and its variants. Naïve clustering performs progressive clustering in the EE stage, while the SAC stage matches that of SAC-ViT. In random clustering, the EE stage aligns with SAC-ViT, but the SAC stage randomly clusters into two classes.}
\label{tab:tab7}
\end{table}

\textbf{Comparison with SOTA ViT Optimization Models.} Table~\ref{tab:tab4} compares SAC-ViT with state-of-the-art (SOTA) ViT optimization methods. SAC-ViT balances accuracy and computational efficiency. At lower thresholds ($\eta = 0.45$), it reduces FLOPs to 1.8G while maintaining a Top-1 accuracy of 79.8\%, comparable to DeiT-S \cite{touvron2021training} (79.8\% with 4.6G FLOPs), DynamicViT \cite{rao2021dynamicvit} (79.3\% with 2.9G FLOPs), and PS-ViT \cite{tang2022patch} (79.9\% with 2.1G FLOPs). At higher thresholds ($\eta = 0.62$), SAC-ViT achieves 80.8\% accuracy with 2.2G FLOPs, surpassing IA-RED$^2$ \cite{pan2021ia} (79.2\% with 3.2G FLOPs), CF-ViT \cite{chen2023cf} (80.0\% with 3.3G FLOPs), and LF-ViT \cite{hu2024lf}. Using a higher resolution ($288 \times 288$), SAC-ViT achieves 82.3\% accuracy with 3.2G FLOPs, outperforming CF-ViT (81.9\% with 4.8G FLOPs) and LF-ViT (82.2\% with 3.7G FLOPs).

SAC-ViT's superior performance is attributed to its non-iterative semantics-aware clustering strategy, which conserves computational resources by allowing simple samples to exit early and reduces the computational burden for difficult samples through local self-attention after clustering. By clustering target and non-target tokens, SAC-ViT minimizes the influence of non-target tokens on the target, thereby enhancing efficiency. This effectiveness is supported by previous methods \cite{chen2023cf,rao2021dynamicvit,meng2022adavit} that have successfully reduced the number of tokens while maintaining or improving accuracy.

\textbf{Comparing with SOTA adaptive ViT optimization methods.} Fig.~\ref{fig:fig1} compares our SAC-ViT with various adaptive ViT optimization methods. The results demonstrate that SAC-ViT not only reduces computational costs but also maintains or improves accuracy, achieving a superior balance between efficiency and accuracy compared to other methods. This balance makes SAC-ViT particularly effective in scenarios where both computational resources and model performance are critical.

\begin{figure}[t]
\centering
\includegraphics[width=1.0\columnwidth]{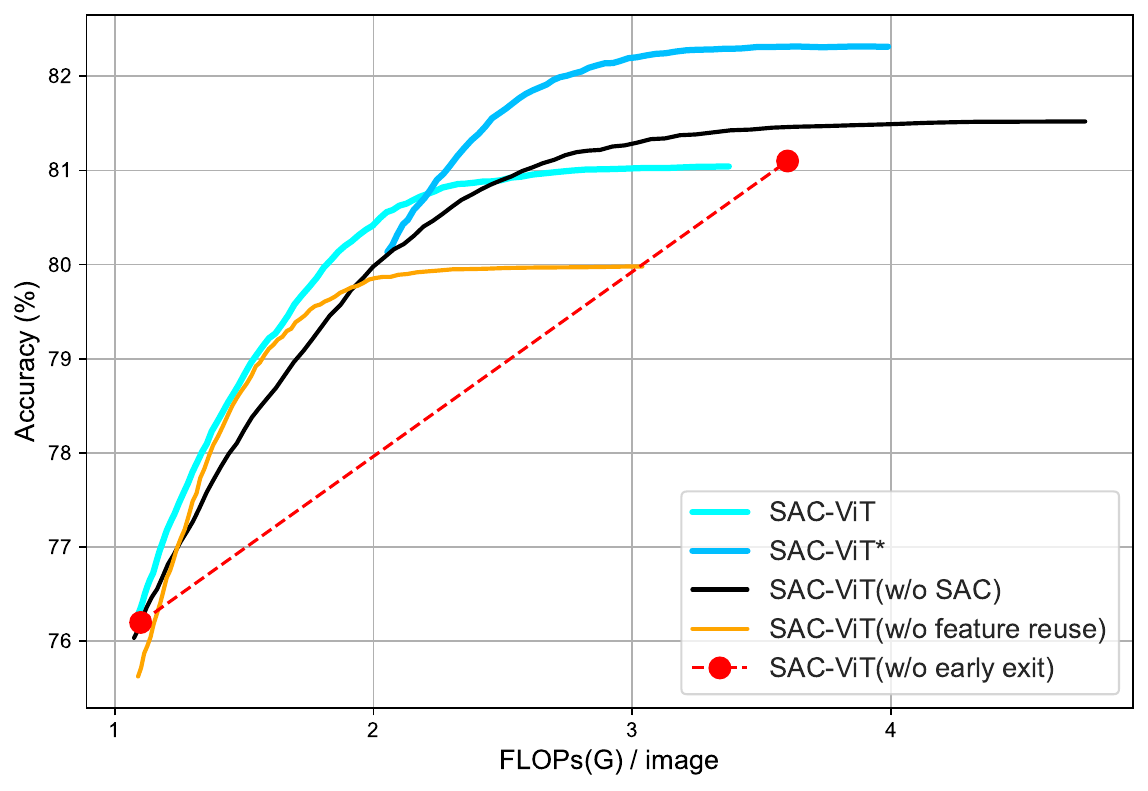}
\caption{Performance analysis of removing each of designs.}
\label{fig:fig3}
\end{figure}

\textbf{Comparison between SAC-ViT and its variants.} Table~\ref{tab:tab7} compares SAC-ViT with its variants.  Naïve Clustering, which progressively clusters images with model depth, and Random Clustering, which randomly selects a token as the clustering center and computes the KL distances of other tokens for clustering. Clustering categories are set to 8, 4, and 2 in the 3rd, 6th, and 9th layers, respectively.

The table shows that SAC-ViT significantly outperforms the naïve clustering method in terms of accuracy. This improvement is mainly due to the naïve approach performing progressive clustering during the low-resolution EE stage before the image's semantic information is accurately extracted. In contrast, our method uses the global moving average of class tokens, which better reflects token importance, thus also surpassing random clustering. This finding aligns with previous research \cite{chen2023cf,hu2024lf} demonstrating that class tokens effectively indicate token importance.


\begin{figure}
\centering
\includegraphics[width=1.0\columnwidth]{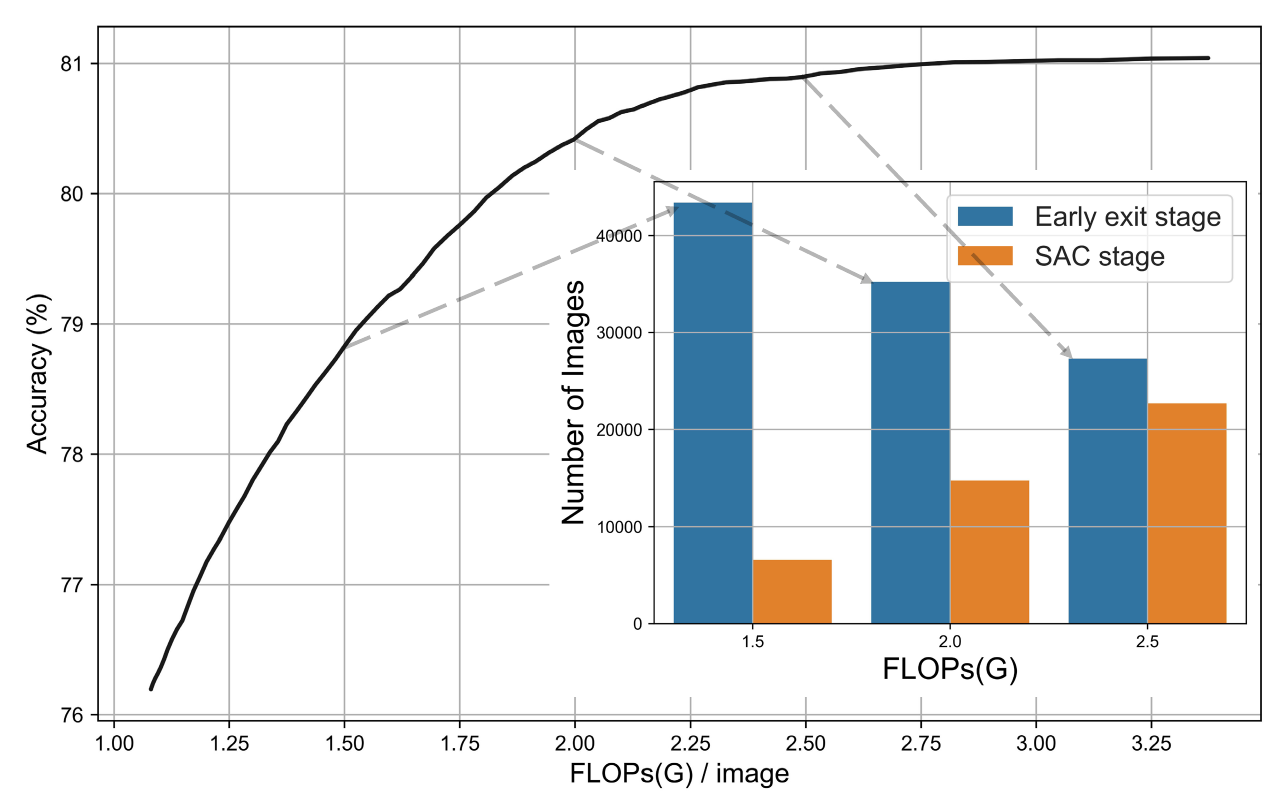}
\caption{Quantitative  analysis of SAC-ViT's early exit stage and semantic-aware clustering stage.}
\label{appedixfig:fig1}
\end{figure}

\begin{figure*}[ht]
\centering
\includegraphics[width=1.8\columnwidth]{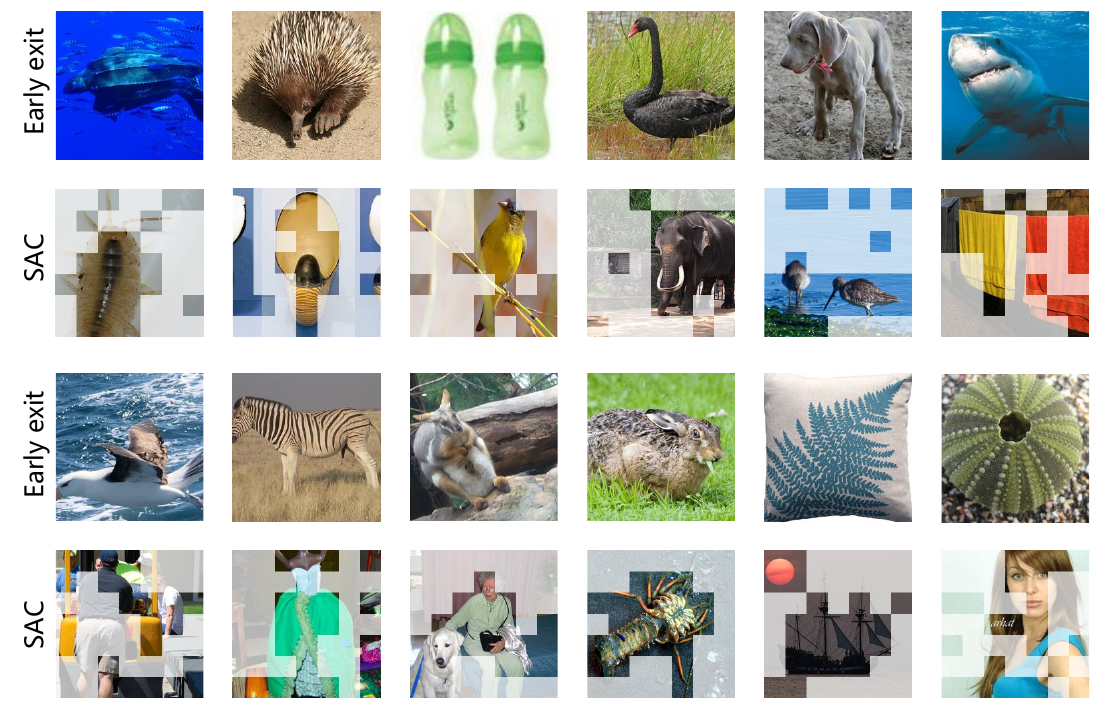}
\caption{Qualitative analysis of SAC-ViT's early exit and semantic-aware clustering stages reveals that images recognized during the early exit stage generally have simple backgrounds, with the objects of interest clearly visible throughout the image. In contrast, during the SAC stage, SAC-ViT correctly identifies objects with complex backgrounds by clustering the target and background based on semantic information, followed by performing local self-attention to enhance recognition efficiency. (The shaded and non-shaded regions of the SAC stage indicate non-target and target clustering, respectively.)}
\label{appedixfig:fig2}
\end{figure*}

\paragraph{Quantitative Analysis.}
Fig.~\ref{appedixfig:fig1} shows the accuracy of SAC-ViT under different computational costs (FLOPs) and the number of images allocated to the Early Exit (EE) stage and the Semantic-Aware Clustering (SAC) stage. Specifically, as the computational cost increases, the number of images processed by the SAC stage rises significantly, directly contributing to an overall improvement in the model's accuracy. The figure clearly demonstrates that with more computational resources, a greater number of complex images are assigned to the SAC stage for processing, allowing the model to make full use of the additional computational power. For example, when the computational cost is 1.5 GFLOPs, 43,000 images are allocated to the EE stage, resulting in an accuracy of 78.8\%. However, as the computational cost increases to 2.5 GFLOPs, the number of images allocated to the EE stage decreases to 28,000, while the overall accuracy rises to 80.8\%. This indicates that SAC-ViT can adaptively adjust the allocation of images between the EE and SAC stages based on different computational budgets, achieving an optimal balance between computational efficiency and accuracy. This flexible allocation mechanism not only ensures high-precision outputs but also enhances the overall computational efficiency of the system. Additionally, as shown in Table 1 of the main paper, this two-stage recognition method significantly reduces spatial redundancy in the images, thereby lowering the demand for computational resources and greatly improving the model's resource utilization during actual deployment. This finding suggests that SAC-ViT has broad application potential in handling large-scale image datasets, as it can effectively conserve computational resources while maintaining high accuracy.

\paragraph{Qualitative analysis.}
In Fig.~\ref{appedixfig:fig2}, we conducted a qualitative analysis of SAC-ViT. By observing the examples in the figure, we found that images recognized in the EE stage typically have simpler backgrounds, with the target object occupying most of the image area and featuring clear and easily distinguishable contours. Such images generally contain less information, allowing them to be quickly and accurately processed in the resource-limited EE stage. In contrast, images processed in the SAC stage are more complex, with rich and cluttered background information, lower contrast between the target and the background, and a relatively smaller proportion of the target object in the image. These complex images require higher computational capacity and more sophisticated model processing, which is why they are allocated to the SAC stage for recognition. This automated allocation strategy not only optimizes the use of computational resources but also ensures high recognition accuracy when dealing with both simple and complex images.

Moreover, this image allocation mechanism reflects SAC-ViT's adaptability and flexibility in handling different types of images. Simple images are quickly processed in the EE stage, reducing computational overhead, while complex images undergo in-depth analysis in the SAC stage to ensure final recognition accuracy. This layered processing model effectively balances computational resources and recognition performance, further illustrating SAC-ViT's practicality and reliability in dealing with diverse image data in real-world applications. Through this mechanism, SAC-ViT not only provides an efficient solution in resource-constrained environments but also demonstrates exceptional performance in handling complex scenarios, thereby enhancing the model's potential for widespread application.

\subsection{Ablation Study}

\begin{table}[t]
\centering
\begin{tabular}{ccccccc }
\hline
    $\beta$ & 0.0 & 0.5 & 0.9 & \textbf{0.99} & 0.999 \\
    \hline
    Acc.(\%) & 80.9 & 81.0 & 81.1 & 81.1 & 81.1 \\
    \hline
\end{tabular}
\caption{Accuracy with different values of $\beta$.}
\label{tab:tab5}
\end{table}

\begin{table}[t]
\centering
\begin{tabular}{ccccccc}
\hline
    $\alpha$ & 0.4 & \textbf{0.5} & 0.6 & 0.7 & 0.8 & 0.9 \\
    \hline
    Acc.(\%) & 80.5 & 81.1 & 81.4 & 81.4 &  81.6 & 81.6\\
    \hline
\end{tabular}
\caption{Accuracy with different values of $\alpha$.}
\label{tab:tab6}
\end{table}

\textbf{Necessity of Each Design.} Fig.~\ref{fig:fig3} plots the performance of our SAC-ViT by individually removing each design component. From the results, we observe that each component of SAC-ViT plays a crucial role in its performance. Feature fusion contributes a 1\% accuracy improvement, while the early exit strategy allows SAC-ViT to adjust accuracy based on computational requirements adaptively.

\textbf{Influence of $\beta$ and $\alpha$.} 
Tables~\ref{tab:tab5} and~\ref{tab:tab6} show the effects of $\beta$ and $\alpha$ on the accuracy of the SAC stage in SAC-ViT, respectively. Here, $\beta$ represents the weight of the shallow encoder in calculating the global moving average attention score, while $\alpha$ denotes the ratio of target tokens to total tokens in the EE phase. In this paper, we choose $\alpha=0.5$ and $\beta=0.99$ by default to achieve the optimal balance between precision and cost.

\textbf{Influence of Loss Function.} During the training of SAC-ViT using Eq.~\ref{eq9}, we use the Cross-Entropy (CE) loss function to supervise the output of the SAC stage with ground truth (GT) labels. For the output of the EE stage, we use the Kullback-Leibler (KL) loss function, supervising it with the output of the SAC stage. To further investigate the impact of different loss functions on SAC-ViT's performance, we also employ the CE loss function and use GT labels to supervise the EE inference outputs of SAC-ViT:
\begin{equation}
    \hat{\mathcal{L}_{cls}} = CE(\mathbf{p}_{sac}; \mathbf{y}) + CE(\mathbf{p}_{ee}; \mathbf{y}),
\end{equation}
Table~\ref{tab:tab8} shows the impact of different loss functions on the performance of SAC-ViT. From the table, we observe that the $CE + KL$ loss improves accuracy by 0.3\% in the EE stage and 0.4\% in the SAC stage compared to the $CE + CE$ loss. Therefore, we select $CE + KL$ as our default loss function. We believe the higher accuracy achieved with $CE + KL$ is due to knowledge distillation transferring the output from the EE stage to the SAC stage, resulting in better performance.

\begin{table}[t]
\centering
\begin{tabular}{ccc}
\hline
     \multirow{2}{*}{  Ablation}  & \multicolumn{2}{c}{ Top-1 Acc.(\%)}\\
    
     & early exit & SAC \\
\hline
CE + CE & 75.9 & 80.7 \\
    \hline
    \textbf{CE + KL(Ours)} & \textbf{76.2} & \textbf{81.1} \\
    \hline
\end{tabular}
\caption{Performance comparison between different loss functions.}
\label{tab:tab8}
\end{table}

\section{Conclusion}
This paper introduces SAC-ViT, an optimization method for Vision Transformers (ViTs) that enhances efficiency through Early Exit (EE) strategies and Semantics-Aware Clustering (SAC) for localized self-attention. SAC-ViT operates in two stages: EE and non-iterative SAC. Initially, the input image is downsampled, and ViT extracts global semantic information at a low cost, generating initial inference. If the results do not meet the EE criteria, the semantic information is clustered into target and non-target groups. Target clusters are mapped back to the original image, cropped, and processed with ViT alongside reused non-target clusters. This two-stage approach reduces spatial redundancy, improves efficiency through localized self-attention, and achieves an excellent balance between accuracy and computational cost.

\bibliography{aaai25}


\end{document}